\documentclass[10pt,twocolumn,letterpaper]{article}

\usepackage{authblk}
\usepackage[multiple]{footmisc}
\usepackage{bigfoot}
\usepackage[accsupp]{axessibility}
\usepackage{amsmath,bm}
\makeatletter
\renewcommand\AB@affilsepx{, \protect\Affilfont}
\makeatother

\usepackage{cvpr}              
\usepackage{blindtext}
\usepackage{multirow}

\usepackage[dvipsnames]{xcolor}

\definecolor{cvprblue}{rgb}{0.21,0.49,0.74}
\usepackage[pagebackref,breaklinks,colorlinks,citecolor=cvprblue]{hyperref}

\newcommand*\samethanks[1][\value{footnote}]{\footnotemark[#1]}

\def\paperID{8764} 
\def\confName{CVPR}
\def\confYear{2024}

\title{PikeLPN: Mitigating Overlooked Inefficiencies of Low-Precision \\ Neural Networks}

\author[1]{Marina Neseem\thanks{Work done during internship at Google.}\thanks{Corresponding authors: marina\_neseem@brown.edu and danielemoro@google.com}}
\author[2]{Conor McCullough}
\author[2]{Randy Hsin}
\author[2]{Chas Leichner}
\author[2]{Shan Li}
\author[2]{In Suk Chong}
\author[2]{Andrew Howard}
\author[2]{Lukasz Lew}
\author[1]{Sherief Reda}
\author[2]{Ville-Mikko Rautio}
\author[2]{Daniele Moro\samethanks}
\affil[1]{Brown University}
\affil[2]{Google}

\begin{document}


\maketitle

\vspace*{-1.2cm}

\begin{abstract}

Low-precision quantization is recognized for its efficacy in neural network optimization.
Our analysis reveals that non-quantized elementwise operations which are prevalent in layers such as parameterized activation functions, batch normalization, and quantization scaling dominate the inference cost of low-precision models.
These non-quantized elementwise operations are commonly overlooked in SOTA efficiency metrics such as Arithmetic Computation Effort (\textit{ACE}) \cite{zhang2022pokebnn}.
In this paper, we propose $ACE_{v2}$ \-- an extended version of \textit{ACE} which offers a better alignment with the inference cost of quantized models and their energy consumption on ML hardware.
Moreover, we introduce \textit{PikeLPN}\footnote{\textbf{Pike} is a slim fast fish, \textbf{LPN} stands for Low-Precision Network.}, a model that addresses these efficiency issues by applying quantization to both elementwise operations and multiply-accumulate operations. In particular, we present a novel quantization technique for batch normalization layers named \textit{QuantNorm} which allows for quantizing the batch normalization parameters without compromising the model performance. 
Additionally, we propose applying \textit{Double Quantization} where the quantization scaling parameters are quantized. 
Furthermore, we recognize and resolve the issue of distribution mismatch in Separable Convolution layers by introducing \textit{Distribution-Heterogeneous Quantization} which enables quantizing them to low-precision.
\textit{PikeLPN} achieves Pareto-optimality in efficiency-accuracy trade-off with up to $3\times$ efficiency improvement compared to SOTA low-precision models.
\end{abstract}    
\section{Introduction}
\label{sec:intro}

\begin{figure}[t]
  \centering
   \includegraphics[width=0.95\linewidth]{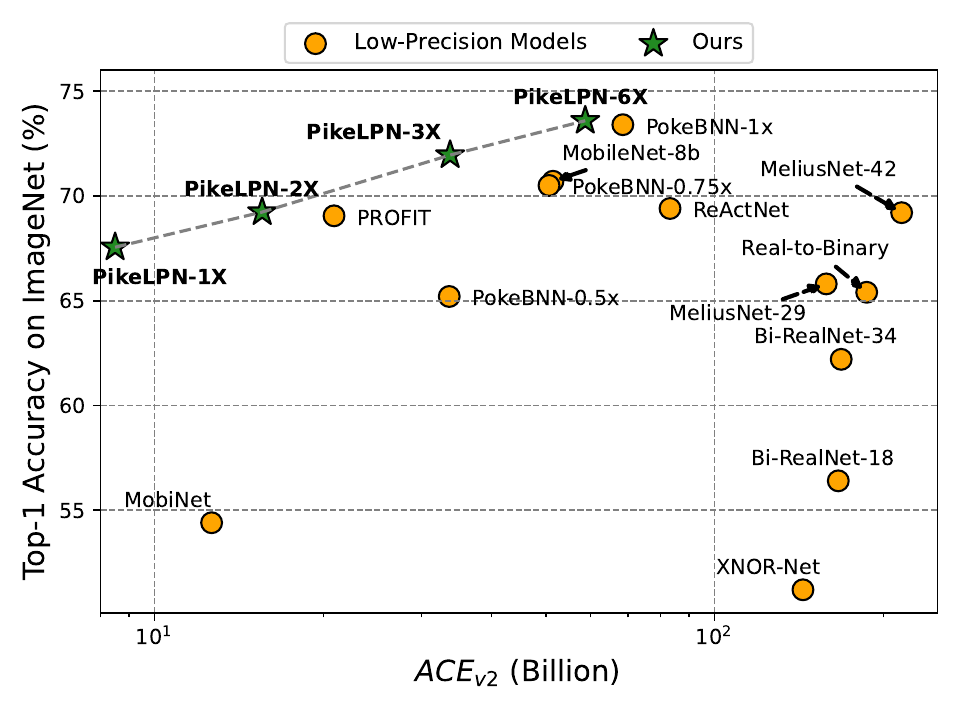}
  \vspace{-10pt}

   \caption{Accuracy vs $ACE_{v2}$ of PikeLPN and SOTA low-precision neural networks.
   $ACE_{v2}$ is an efficiency metric that estimates the cost of arithmetic operations during inference.}
  \vspace{-15pt}
   \label{fig:sota_ace}
\end{figure}

\begin{figure}[t]
  \centering
   \includegraphics[width=0.95\linewidth]{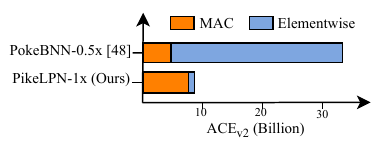}
   \vspace{-15pt}
   \caption{Contribution of multiply-accumulate (MAC) versus elementwise operations to the commonly used efficiency metric $ACE_{v2}$ for PikeLPN-1X and PokeBNN-0.5X \cite{zhang2022pokebnn}.
   \textit{PikeLPN} selectively increases the precision of MAC operations which allows for effectively quantizing elementwise operations, achieving $3\times$ more efficiency while being 2\% more accurate on ImageNet.}
   \vspace{-15pt}
   \label{fig:macs_vs_ele_example}
\end{figure}

Quantization has long been established as a method to decrease the precision of neural network weights and activations effectively, resulting in smaller models and accelerated processing \cite{gholami2022survey}. 
Recent studies have shown impressive results in image classification tasks, making the use of low-precision quantization (i.e., 4 bits or fewer) increasingly popular \cite{park2020profit, zhang2022pokebnn, liu2020reactnet, phan2020mobinet}.
In these compact models, convolutional and fully connected layers are typically constrained to 4-bit precision or even less, while precision is maintained at higher levels in other layers of the network. 
For example, the state-of-the-art (SOTA) binary network PokeBNN \cite{zhang2022pokebnn} binarizes the convolutional layers of ResNet-50 \cite{he2016deep}, and to avoid accuracy loss, they incorporate extra skip connections, extra batch normalization layers, and parameterized activation functions (DPReLU) that are executed in high precision.
As illustrated in Figure \ref{fig:macs_vs_ele_example}, while this strategy significantly reduces the cost of multiply-accumulate (MAC) operations, it shifts the energy burden to the elementwise operations within these remaining high-precision layers.
Although there are fewer of these elementwise operations, they use more energy because they are still in high precision.
This indicates a critical area of optimization to improve the overall efficiency of low-precision models.

We analyze the key efficiency bottlenecks in low-precision models uncovering a fundamental limitation of the efficiency metrics in literature, ACE \cite{zhang2022pokebnn}, CPU64 \cite{real_to_binary, liu2020reactnet}, Unit-gate model \cite{zimmermann1999computer} and FA-count \cite{sakr2017analytical}.
Those metrics exclude the elementwise operations in arithmetic calculations, a sentiment grounded in the belief that their contribution to the total computation cost is negligible compared to MAC operations.
Optimizing for those metrics drives researchers to prioritize the reduction of computational precision in Convolutional and Dense layers, yet they overlook the quantization of elementwise operations.
As a result, operations such as batch normalization, activation functions, and quantization scaling multiplications, are often performed at full precision.
Moreover, SOTA low-precision models tend to rely extensively on mechanisms like branching \cite{huang2017densely} and skip connections \cite{he2016deep}, which significantly increase energy costs associated with memory reads and writes.
To overcome this issue, we propose $ACE_{v2}$ which extends the efficiency metric \textit{ACE} to account for all arithmetic operations in quantized neural networks including both elementwise and MAC operations.
This would help guide researchers' choices when designing low-precision models.

Guided by our $ACE_{v2}$ metric, we design \textit{PikeLPN} \--- a novel family of efficient low-precision models. \textit{PikeLPN} quantizes both elementwise and MAC operations. 
Remarkably, \textit{PikeLPN} not only achieves a 3$\times$ cost reduction compared to SOTA binary models \cite{liu2020reactnet, zhang2022pokebnn}, it also achieves competitive accuracy levels on ImageNet \cite{imagenet}.

\noindent \textbf{Our contributions} can be summarized as follows:
\begin{itemize}
    \item We identify and analyze the overlooked cost of non-quantized elementwise operations in SOTA low-precision models.
    Our analysis shows that the non-quantized elementwise operations used in parameterized activation functions, batch normalization, and quantization scaling dominate the inference cost of low-precision models.
    \item We propose $ACE_{v2}$ \--- an extension to the existing hardware-agnostic cost metric \textit{ACE}. $ACE_{v2}$ offers a better alignment with the cost of the low-precision models and their energy consumption on ML hardware by accounting for all arithmetic operations during inference.
    \item We propose \textit{PikeLPN} \--- a novel family of low-precision architectures, which improves the efficiency of low-precision models by quantizing both elementwise and multiply-accumulate operations. 
    Specifically, we propose (a) \textit{QuantNorm} for effective batch normalization quantization, (b) \textit{Double Quantization} where quantization parameters are also quantized, and (c) \textit{Distribution-Heterogeneous Quantization} for Separable Convolution layers to tackle their distribution mismatch problem.
\end{itemize}

The rest of the paper is organized as follows. We review the related work in Section~\ref{sec:related_work}. In Section~\ref{sec:methood}, we propose $ACE_{v2}$ providing detailed analysis to the overlooked efficiency bottlenecks by previous cost metrics.
Then, guided by the new cost metric, we propose our efficient \textit{PikeLPN} model.
Next, we compare PikeLPN to SOTA low-precision models in Section~\ref{sec:experiments}. 
Finally, we conclude in Section~\ref{conclusion}.
\section{Related Work}
\label{sec:related_work}

\textbf{Low-precision Quantization:} A substantial body of work exists in the realm of low-precision quantization, exemplified by studies that indicate that architectures can be quantized to 4 bits with minimal impact on accuracy \cite{lowprecisionquantization, jung2019learning, abdolrashidi2021pareto, park2020profit}.
Others perform logarithmic quantization methods known for their hardware efficiency \cite{tann_multiplier_free,hashemi2017understanding, li2019additive}.
In addition, there are attempts to push the boundaries by introducing predominantly binary models where some of the convolution layers are quantized to 1 bit while other layers are maintained at a higher precision \cite{zhang2022pokebnn, liu2020reactnet, bin_mobilenet}. 
Some researchers have also developed automated strategies for mixed-precision modeling to dynamically choose the optimal precision for each layer, contingent upon a predetermined efficiency metric \cite{oneshotquant2023}. 
However, existing approaches primarily focus on the quantization of multiply-accumulate (MAC) operations in convolution and dense layers. 
They commonly neglect elementwise operations such as those in batch normalization layers and activation functions. 
Our empirical findings show that this assumption becomes invalid for low-precision models, specifically 4 bits or below. 

\textbf{Architectural Approaches to Low-precision Models:}
Several studies have adopted architectural modifications to enhance the performance of low-precision models.
Many such modifications involve the integration of modules consisting solely of elementwise operations, aiming to minimize computational and parameter overhead.
For instance, the channelwise real-valued rescaling of binarized tensors has been proposed as an effective means to reduce quantization error \cite{rastegari2016xnor}. 
This approach incorporates elementwise floating-point multiplications for each channel. 
Additional methods, as suggested in \cite{dai2021vs}, advocate for per-vector quantization, which results in multiple elementwise multiplications per channel. 
Studies like FracBNN \cite{zhang2021fracbnn} and PokeBNN \cite{zhang2022pokebnn} include extra Batch Normalization layers in their predominantly binary models to expedite the training convergence.
Moreover, the use of parameterized activation functions, such as PReLU \cite{prelu} and DPReLU \cite{zhang2022pokebnn}, has become a standard practice for improving the performance of low-precision models \cite{liu2020reactnet, liu2021adam}.
All these modifications necessitate elementwise floating-point multiplications and additions.
Moreover, the introduction of skip connections has proven beneficial in enhancing low-precision model quality.
Notably, ReActNet \cite{liu2020reactnet} and PokeBNN \cite{zhang2022pokebnn} are designed with 4 and 3 parallel branches, respectively.
Although skip connections only involve elementwise additions, they contribute to an increased memory access during inference to store multiple activations increasing the inference cost \cite{arithmetic_intensity}.

\textbf{Cost Metrics for Efficiency Evaluation:}
MAC operations have been recognized in literature as the principal contributors to inference cost of deep learning models. 
As a result, efficiency metrics have predominantly focused on these specific operations. 
The \textit{CPU64} metric \cite{liu2021adam, liu2020reactnet, liu2018bi} has been used to gauge the efficiency of mixed-precision neural networks when running on CPUs.
With the growing utilization of specialized machine learning hardware and accelerators, a newer metric named \textit{ACE} has been introduced \cite{zhang2022pokebnn}. 
\textit{ACE}, an acronym for \textit{Arithmetic Computation Effort}, is formulated as the product of the number of MAC operations and the bitwidth of the two operands involved, which is directly proportional to the number of active hardware bit-adders required.
The Unit-gate model \cite{zimmermann1999computer} and FA-count \cite{sakr2017analytical} correlate very well with ACE and differ only by a small constant factor \footnote{They do not account for carry-save format for local accumulator representations typically used in systollic arrays.}.
All these metrics do not consider elementwise operations. 
Thus, in this paper, we extend the ACE metric introducing $ACE_{v2}$, and this extension should generalize to other metrics as well.
All these metrics, including the extended ACE, are technology node independent.
\section{Method}
\label{sec:methood}
In this section, we identify previously overlooked costs in state-of-the-art (SOTA) cost metrics. Additionally, we propose extending the \textit{Arithmetic Computational Effort} (ACE) metric \cite{zhang2022pokebnn} to provide a more accurate representation of the inference cost of low-precision models.
Subsequently, we assess the impact of various design alternatives in low-precision models on the cost of inference.
Finally, we present \textit{PikeLPN} \--- a novel family of low-precision models.

\begin{figure}[t]
  \centering
   \includegraphics[width=\linewidth]{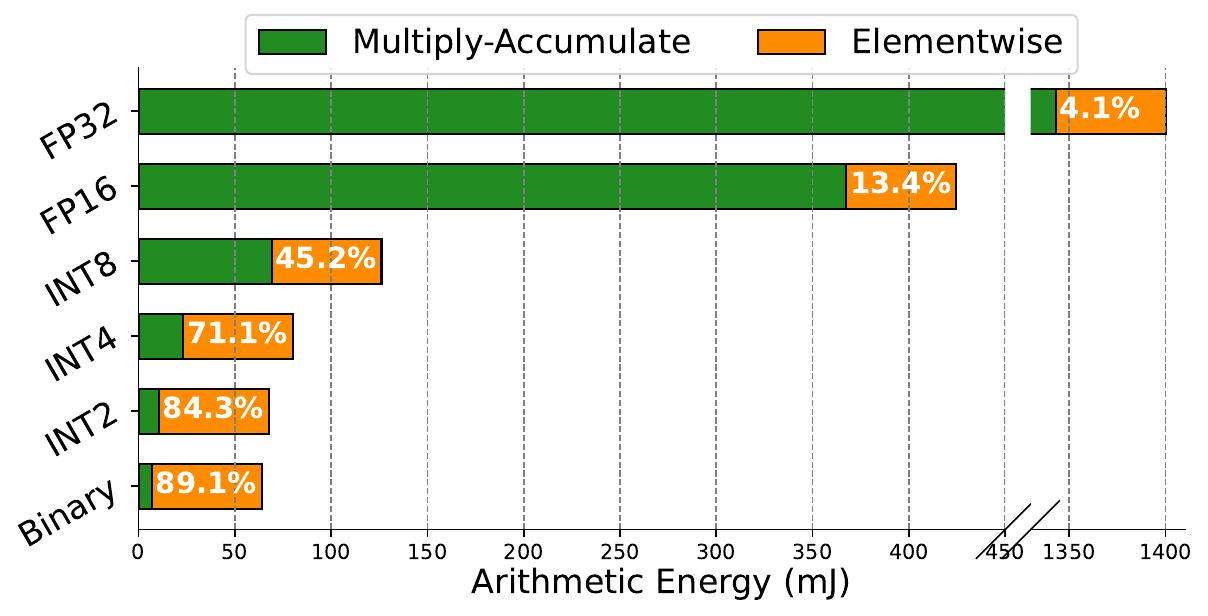}
   \caption{Arithmetic Energy on 45nm CMOS technology by multiply-accumulate operations versus non-quantized elementwise operations for MobileNetV2. Energy costs are calculated using Table \ref{tab:cost_metrics}.
   The figure reveals that elementwise operations are a substantial contributor to the overall cost in low-precision models.}
   \vspace{-15pt}

   \label{fig:macs_vs_ele}
\end{figure}

\subsection{Cost Metrics for Low Precision Models}
The prevalent notion is that multiply-accumulate operations in the convolution and dense layers are the sole substantial contributors to inference cost in deep learning models \cite{park2020profit, zhang2022pokebnn, liu2020reactnet}.
This viewpoint stems from the observation that for full precision models the energy cost of those layers is more than 95\% of the total model operations as shown in Figure \ref{fig:macs_vs_ele}.
Consequently, commonly used efficiency metrics for quantized neural networks, such as \textit{CPU64} \cite{liu2021adam, liu2020reactnet, liu2018bi} and \textit{ACE} \cite{zhang2022pokebnn}, are tailored to exclusively account for multiply-accumulate operations in these specified layers. 
Optimization in accordance with these metrics drive researchers to prioritize reducing the precision of multiply-accumulate operations in convolution and dense layers while maintaining high precision for all other elementwise operations.
Moreover, they re-parameterize the models adding layers that only have elementwise operations to compensate for any accuracy losses by low-precision quantization \cite{zhang2022pokebnn, liu2020reactnet}.
However, our analysis reveals that these non-quantized elementwise operations substantially contributes to the arithmetic cost during inference of low-precision models (i.e., 8 bits and lower), thereby challenging the prevailing assumptions.

Figure \ref{fig:macs_vs_ele} illustrates the relative contributions of low-precision multiply-accumulate operations and non-quantized elementwise operations to the total energy consumption by arithmetic computations at various precisions.
The data reveals a notable trend: the proportion of energy consumed by elementwise operations becomes more significant as the precision decreases. 
For example, in binary-quantized models, those non-quantized elementwise operations account for up to 89\% of the total cost. 
This observation highlights the limitations of existing metrics in accurately gauging the efficiency of quantized models.
Consequently, we propose $ACE_{v2}$ which extends the  \textit{ACE} metric \cite{zhang2022pokebnn} to account for both multiply-accumulate operations as well as elementwise operations.
We anticipate that our comprehensive $ACE_{v2}$ metric will enable more informed optimization choices within the research community.

\subsection{Introducing \texorpdfstring{$\bm{ACE_{v2}}$}{ACEv2}}
\label{extending_ace}
\textit{ACE} has been used to estimate the cost of inference on idealized ML hardware implemented with CMOS methodology \cite{zhang2022pokebnn}. 
\textit{ACE} is defined by its authors as
the number of bitadders (i.e., digital circuit adding 3 bits to form a 2 bit number \--- carry and sum) required to perform every multiply-accumulate operation.
The authors justify that definition by showing a high correlation coefficient (i.e., 0.946) between the number of bitadders and the independently measured energy consumption on 45nm CMOS technology.
While \textit{ACE} provides a hardware-agnostic method to evaluate the efficiency of quantized neural networks, it fails to include the elementwise operations which can be the dominating cost factor in low precision models as shown in Figure \ref{fig:macs_vs_ele}.
Moreover, \textit{ACE} does not provide a way to estimate the cost of shift operations which are required to implement non-linear base-2 logarithmic quantization \cite{you2020shiftaddnet, you2023shiftaddvit}.
We propose $ACE_{v2}$ which improves \textit{ACE} by extending it to include elementwise multiplication, elementwise addition, and shift operations. 
We establish the $ACE_{v2}$ formulas for the previously discussed operations as shown in Table \ref{tab:cost_metrics}. 
\vspace{-10pt}
\begin{table}[t]
  \centering
  \setlength{\tabcolsep}{2pt}
\renewcommand{\arraystretch}{0.8}
  \caption{Cost under 45nm CMOS technology \cite{you2023shiftaddvit, Horowitz_energy} \protect\footnotemark. 
  $f(i,j)$ refers to the formula used to calculate the $ACE_{v2}$ cost where $i$ and $j$ are the precisions of the two operands. $c_a = 6$ and $c_s = 5$.
  The correlation coefficient between $ACE_{v2}$ and the independently measured arithmetic energy consumption is 0.991.
  }
    \vspace{-5pt}
\small
  \begin{tabular}{c | c c | c c | c c }
    \toprule
    & \multicolumn{2}{c}{\textbf{MULTIPLY}} & \multicolumn{2}{c}{\textbf{ADD}} & \multicolumn{2}{c}{\textbf{SHIFT}} \\
    \midrule
    & \textbf{Energy} & \multirow{2}{*}{$\bm{ACE_{v2}}$} & \textbf{Energy} & \multirow{2}{*}{$\bm{ACE_{v2}}$} & \textbf{Energy} & \multirow{2}{*}{$\bm{ACE_{v2}}$} \\
    & ($pJ$) & & ($pJ$) & & ($pJ$) & \\
    \midrule
    FP32 & 3.7 & 992 & 0.9 & 192 & - & - \\
    FP16 & 1.1 & 240 & 0.4 & 96 & - & - \\
    \midrule
    $f(i,j)$ & \multicolumn{2}{c}{$i \cdot j$ - $max(i, j)$} & \multicolumn{2}{c}{$c_a \cdot max(i, j)$} & \multicolumn{2}{c}{-}\\
    \midrule
    INT32 & 3.1 & 992 & 0.1 & 32 & 0.13 & 32 \\
    INT16 & - & 240 & - & 16 & 0.057 & 12.8 \\
    INT8 & 0.2 & 56 & 0.03 & 8 & 0.024 & 4.8 \\
    INT4 & - & 12 & - & 4 & - & 1.6 \\
    INT2 & - & 2 & - & 2 & - & 0.4 \\
    Binary & - & - & - & 1 & - & - \\
    \midrule
    $f(i,j)$ & \multicolumn{2}{c}{$i \cdot j$ - $max(i, j)$} & \multicolumn{2}{c}{$max(i, j)$} & \multicolumn{2}{c}{$i\cdot log_2(j)/c_s$}\\
    \bottomrule
  \end{tabular}
  \vspace{-15pt}
  \label{tab:cost_metrics}
\end{table}

\footnotetext{Energy costs for low-precision operations can be extrapolated linearly for addition and quadratically for multiplication \cite{wallace_and_dadda}.}

\noindent \textbf{Elementwise Multiplications:} 
Using established methods for constructing multipliers, such as adder trees proposed by Wallace and Dadda \cite{Wallace, Dadda}, we calculated the number of adders needed to multiply an $i$-bit number by a $j$-bit number as $i \cdot j - max(i, j)$. This formula exactly matches the optimal number of adders for $1 <= i, j <= 64$. See Section \textcolor{red}{6} in the Appendix for a detailed explanation. 


\noindent \textbf{Elementwise Additions:}
Fixed-point numbers added using established adders \footnote{While there are many methods for constructing adders, such as Carry Lookahead Adder \cite{carry_lookahead_adder} and Ripple Carry Adder \cite{rca}, the particular implementation has a limited effect on the energy use.} activate an upper bound of $max(i, j)$ bit adders to add i-bit and j-bit numbers.
Floating-point adders additionally require exponent alignment, significand addition, and normalization steps \cite{fp_adders}, resulting in a much higher energy consumption compared to fixed-point adders as shown in Table \ref{tab:cost_metrics}. 
We analyze the operations needed in floating point adders \cite{fp_adders} and come to an $ACE_{v2}$ cost of $6\times$ the cost of a fixed-point adder. Therefore, we derive $ACE_{v2}$ for floating point adders using $c_a \cdot max(i, j)$ with $c_a=6$.
See Appendix Section \textcolor{red}{7} for a detailed explanation.

\noindent \textbf{Shift Operations:} 
A Barrel Shifter is an established method to shift and rotate $i$-bit numbers by $j$ locations in modern processors \cite{barrel_shifter}.
The barrel shifter is implemented as a cascade of $i\log_2(j)$ \textit{2$:$1} multiplexers.
Therefore, we derive $ACE_{v2}$ for a shift operation as $i\log_2(j)/c_s$ where $c_s$ is the ratio of the cost of a \textit{2$:$1} multiplexer compared to a full adder.
Since a full adder can be efficiently implemented using five \textit{2:1} multiplexers based on \cite{Journals2015PowerOM}, we assign $c_s=5$.

To verify the correctness of our $ACE_{v2}$ metric, Table \ref{tab:cost_metrics} shows a $0.991$ correlation coefficient between the \textit{independently} measured energy consumption of various arithmetic units on the 45nm CMOS technology and its $ACE_{v2}$ cost, a notable improvement compared to the $0.946$ correlation coefficient in $ACE$ \cite{zhang2022pokebnn}.
Using those definitions, we estimate a more accurate arithmetic cost for any quantized model. 

\subsection{Overlooked Efficiency Bottlenecks}
\label{key_bottlenecks}
\begin{table}[t]
  \centering
  \setlength{\tabcolsep}{2.2pt}
\renewcommand{\arraystretch}{0.8}
  \caption{The contribution of non-quantized Batch Normalization Layers to the overall $ACE_{v2}$ cost.}
    \vspace{-5pt}
\small
  \begin{tabular}{c | c | c | c}
    \toprule
    \multirow{2}{*}{\textbf{Model}} & \textbf{BN Adds} & \textbf{BN Mults} & \textbf{BN $\bm{ACE_{v2}}$} \\
    & $(Million)$ & $(Million)$ & (\%) \\
    \midrule
    MobileNetV2 $(4W,4A)$ & 6.67 & 6.67 & 
    \textbf{41.87} \\
    ResNet50 $(1W,1A)$ & 10.58 & 10.58 & 
    \textbf{41.38} \\
    \bottomrule
  \end{tabular}
  \vspace{-10pt}
  \label{tab:batchnorm_energy}
\end{table}

\noindent \textbf{Batch Normalization:}
Batch normalization layers, which necessitate elementwise multiplications and additions, typically retain parameters in floating-point format during deep neural network quantization to maintain training stability and prevent accuracy loss \cite{zhang2022pokebnn, liu2020reactnet, bin_mobilenet}.
Consequently, these operations are performed using floating-point (FP32) arithmetic, with a single FP32 operation consuming approximately 18$\times$ more energy than an INT8 multiplication, as detailed in Table \ref{tab:cost_metrics}. 
Assessing the impact of these non-quantized batch normalization layers in Table \ref{tab:batchnorm_energy} reveals that they can account for as much as 42\% of the total $ACE_{v2}$ cost in various low-precision models. 
This substantial contribution shows the importance of considering the cost of these operations and potentially quantizing its parameters.

\noindent \textbf{Activation Layers:}
In recent literature, low-precision models have increasingly replaced ReLU \cite{relu} activation functions with parameterized activation functions such as PReLU \cite{prelu} and DPReLU \cite{zhang2022pokebnn} to improve performance and training stability of quantized models \cite{liu2020reactnet, phan2020mobinet}.
The dynamic parameterized rectified linear unit (DPReLU), for instance, is defined by the following piecewise function:
\vspace{-5pt}
\begin{equation}
    DPReLU(x)= 
    \begin{cases}
        \eta(x-\alpha)-\beta & \text{if } x-\alpha > 0\\
        \gamma(x-\alpha)-\beta & \text{otherwise}
    \end{cases}
\vspace{-5pt}
\end{equation}
Here, the parameters $\eta$, $\alpha$, $\beta$, and $\gamma$ are represented in floating-point format.
Consequently, the computation of DPReLU necessitates both elementwise floating-point multiplications and additions. 
Our study, detailed in Table \ref{tab:activation_energy}, assesses the impact of these elementwise operations on the $ACE_{v2}$ cost. 
We find that in a 4-bit MobileNetV2 model, the incorporation of different activation functions — namely ReLU, PReLU, and DPReLU — significantly influences the cost. 
Specifically, the use of PReLU and DPReLU, despite their benefits on accuracy, introduces up to 35\% increase in the overall inference cost. 
This finding highlights the need to balance the benefits of parameterized activation functions with their computational demands.

\begin{table}[t]
  \centering
  \setlength{\tabcolsep}{3.2pt}
\renewcommand{\arraystretch}{0.8}
  \caption{
  The contribution of non-quantized parameterized activation functions to the overall $ACE_{v2}$ cost. Analysis performed by applying different activation functions to a 4-bit MobileNetv2.}
  \vspace{-5pt}
\small
  \begin{tabular}{c | c c | c | c}
    \toprule
    \multirow{2}{*}{\textbf{Activation}} & \textbf{Adds} & \textbf{Mults} & $\bm{ACE_{v2}}$ & \textbf{Overhead}\\
    & $(Million)$ & $(Million)$ &  $(\times 10^9)$ & $(\%)$ \\
    \midrule
    ReLU \cite{relu} & 0 & 0 & 20.44 & - \\
    PReLU \cite{prelu} & 0 & 6.1 & 26.5 & \textbf{+29.6\%} \\
    DPReLU \cite{zhang2022pokebnn} & 6.1 & 6.1 & 27.67 & \textbf{+35.3\%} \\
    \bottomrule
  \end{tabular}
  \vspace{-15pt}
  \label{tab:activation_energy}
\end{table}

\noindent \textbf{Skip Connections:}
Skip connections are regarded as zero-cost operations in terms of arithmetic computation.
Consequently, previous work overused them to improve the model performance without having any measurable effect on the cost \cite{zhang2022pokebnn, liu2020reactnet, bin_mobilenet}.
For instance, ReActNet \cite{liu2020reactnet} incorporated four parallel branches, quadrupling its memory footprint compared to a single-path model. 
PokeBNN \cite{zhang2022pokebnn} followed a similar design, incorporating three parallel branches.
However, such branching necessitates the concatenation of feature maps from previous layers, leading to an increase in the amount of data concurrently stored in memory.
That increase the required memory reads and writes which have significant costs.
As an example, in a processor with a 32KB cache designed using 45nm CMOS technology, moving an 8-bit element from the cache consumes approximately $2.5 pJ$ of energy. This is about $12 \times$ the energy needed for an INT8 multiplication operation, which requires only around $0.2 pJ$ as shown in Table \ref{tab:cost_metrics}. 
This disparity becomes even more profound when data must be transfered from DRAM, where the energy requirement balloon to $162.5 pJ$ \--- $810\times$ higher than the INT8 multiplication \cite{Horowitz_energy}.
Quantifying this overhead in a hardware-agnostic manner is challenging since it is influenced by a multitude of factors including the underlying hardware architecture, memory location, and model size. 
Yet, understanding its impact remains crucial to design efficient models. 
We advocate for the adoption of \textit{Arithmetic Intensity} as a practical metric to measure memory reads and writes during inference \cite{arithmetic_intensity}.
Arithmetic Intensity ($AI_c$) is defined as the ratio of the arithmetic operations ($M_c$) to the amount of data, including both Weights ($W$) and Activations ($A$), required to execute these operations as shown in Equation \ref{eq:cum_arithmetic_intensity}. 
\begin{equation}
    AI_c = \frac{M_c}{W + A}
\label{eq:cum_arithmetic_intensity}
\end{equation}

Consequently, Arithmetic Intensity serves as an indicator of the amount of memory reads and writes to perform computational operations.
Adding branches lead to a substantial increase in the amount of data that must be loaded to execute a relatively small number of operations; hence decreasing the arithmetic intensity as shown in Table \ref{tab:arithmetic_intensity}.

\begin{table}[t]
  \centering
  \setlength{\tabcolsep}{10pt}
\renewcommand{\arraystretch}{0.5}
  \caption{Arithmetic Intensity computed according to Equation (3) for a ResNet-50 model with various number of branches.}
  \vspace{-5pt}

\small
  \begin{tabular}{c c c}
    \toprule
    \multicolumn{3}{c}{\textbf{Arithmetic Intensity} (Ops/Element $\uparrow$)} \\
    \midrule
    2 Branches & 3 Branches & 4 Branches \\
    \midrule
    \textbf{73.5} & 49.66 & 36.75 \\
    \bottomrule
  \end{tabular}
  \vspace{-5pt}
  \label{tab:arithmetic_intensity}
\end{table}

\begin{table}[t]
  \centering
  \setlength{\tabcolsep}{3.2pt}
\renewcommand{\arraystretch}{0.8}
  \caption{
  $ACE_{v2}$ of a 4-bit MobileNetV2 and a binary ResNet50 model with various quantization granularities.
  The \textit{Overhead} represents the percentage of cost required by the extra FP operations due to quantization (i.e. quantization scaling).}
  \small
  \vspace{-5pt}

  \begin{tabular}{c | c | c | c }
    \toprule
    \textbf{Quantization} & \textbf{Mults} & \multicolumn{2}{c}{$\bm{ACE_{v2}}$ $(\times 10^9)\downarrow$} \\
    \textbf{Granularity} & $(Million)$ & Total & Overhead (\%) \\
    \midrule
    \multicolumn{4}{c}{\textbf{MobileNetV2 - $<4W,4A>$}} \\
    \midrule
    Layerwise \cite{gholami2022survey} & 6.67 & 20.44 & \textbf{32.52\%} \\
    Channelwise \cite{gholami2022survey} & 6.67 & 20.44 & \textbf{32.52\%} \\
    Sub-Channelwise \cite{dai2021vs} & 13.35 & 27.06 & \textbf{48.97\%} \\
    \midrule
    \multicolumn{4}{c}{\textbf{ResNet50 - $<1W,1A>$}} \\
    \midrule
    Layerwise \cite{gholami2022survey} & 10.63 & 28.13 & \textbf{32.03\%} \\
    Channelwise \cite{gholami2022survey} & 10.63 & 28.13 & \textbf{32.03\%} \\
    Sub-Channelwise \cite{dai2021vs} & 32.75 & 50.08 & \textbf{63.55\%} \\
    \bottomrule
  \end{tabular}
  \vspace{-15pt}
  \label{tab:quantization_granularity}
\end{table}

\noindent \textbf{Quantization Granularity Overhead:}
Uniform quantization, a widely adopted technique in SOTA low-precision models \cite{bin_mobilenet, park2020profit, zhang2022pokebnn}, transforms discrete integer values, $q$, into continuous real values, $r$  through the affine relation
\vspace{-5pt}
\begin{equation}
    r=S(q-Z)
\vspace{-5pt}
\label{eq:linear_quantizer}
\end{equation}
where $S$ is a scale factor.
$S$ is a critical component of quantization which is typically learned as an arbitrary floating-point value during training. In the inference phase, this necessitates an elementwise multiplication by $S$, contributing to computational overhead \cite{jacob2018quantization}. 
Proper scaling is crucial in quantization to mitigate quantization error enabling quantized models to maintain high accuracy.
Quantization granularity dictates the level at which scaling factors are applied in a model \cite{gholami2022survey}. 
For example, Layerwise quantization assigns a single scale factor based on all weights within a layer. 
Channelwise quantization, widely adopted in state-of-the-art low-precision models, allocates a unique scaling factor to each channel, catering to the varying distributions of weights and potentially enhancing model accuracy. 
Sub-Channelwise quantization takes this further by assigning several scaling factors within each channel, allowing for even finer adjustments at the expense of increased computational cost \cite{dai2021vs}.
All quantization granularities add one or more elementwise multiplications per channel.
Table \ref{tab:quantization_granularity} compares the $ACE_{v2}$ cost of such quantization granularities. 
In the popular Channelwise quantization, the overhead from elementwise multiplications is 32\% of the total cost.

\begin{figure}[t]
  \centering
   \includegraphics[width=0.7\linewidth]{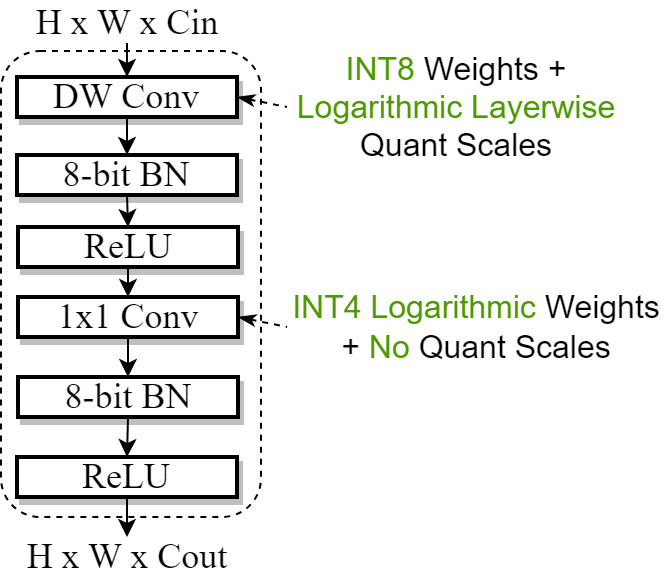}
   \caption{\textit{PikeLPN} building block architecture.
   }
   \label{fig:PikeLPN_block}
   \vspace{-5pt}
\end{figure}

\subsection{PikeLPN Architecture}
\label{pikelpn}
Based on our comprehensive analysis, we introduce \textit{PikeLPN}, a novel architecture engineered to mitigate the inefficiencies of SOTA low-precision models. 
This section introduces the basic block of our proposed \textit{PikeLPN} model, explores quantization strategies for the different layers, and proposes a novel method for quantizing batch normalization layers without compromising the model's accuracy.

\noindent \textbf{PikeLPN Basic Block:}
To engineer an effective low-precision model, we first design the baseline architecture with building blocks that are inherently efficient.
With this principle in mind, our architecture adopts separable convolutional layers, subdivided into depthwise and pointwise convolutions, in line with the framework established by MobileNetV1 \cite{howard2017mobilenets}. 
Those layers are widely recognized for their computational efficiency and have been integrated into SOTA efficient ConvNets \cite{tan2019efficientnet, vasu2023mobileone}.
Figure \ref{fig:PikeLPN_block} illustrates the building block for \textit{PikeLPN}.
To maximize computational efficiency, the used architecture deliberately avoids parameterized activation functions and skip connections that are likely to increase computational cost as explained in Subsection \ref{key_bottlenecks}. 
Finally, our model uses the first and last blocks from the MobileNetV1 architecture due to their proven effectiveness and reliability.

\begin{table}[t]
  \centering
  \setlength{\tabcolsep}{3.2pt}
\renewcommand{\arraystretch}{0.8}
  \caption{Top-1 Accuracy on ImageNet vs $ACE_{v2}$ cost of PikeLPN using various quantizers for the Depthwise and Pointwise Layers. 
  PW-Convolution layers contribute to 95\% of the number of multiply-accumulate operations in the model, that is why we lower the precision of the PW Conv layers to 4 bits while we keep the DW Conv layers at 8-bits.
  }
  \vspace{-5pt}
  \small
  \begin{tabular}{c c | c c | c | c}
    \toprule
    \multicolumn{2}{c}{\textbf{Pointwise Conv.}} & \multicolumn{2}{c}{\textbf{Depthwise Convs}} & \textbf{Top-1} & $\bm{ACE_{v2}}$ \\
    Weights & Q-Params & Weights & Q-Params & (\%) & ($\times 10^9$) \\
    \midrule
    Linear-4 & Arbitrary & Linear-8 & Arbitrary & 68.50 & 20.91 \\
    Linear-4 & PoT & Linear-8 & PoT & 68.41 & 15.93 \\
    \midrule
    PoT-4 & - & PoT-8 & - & 64.50 & 10.05 \\
    \midrule
    PoT-4 & - & Linear-8 & Arbitrary & 67.60 & 12.86 \\
    PoT-4 & - & Linear-8 & PoT &  \textbf{67.55} & \textbf{10.95} \\
    \bottomrule
  \end{tabular}
  \vspace{-15pt}
  \label{tab:quantizers_accuracy_vs_energy}
\end{table}

\begin{figure}[t]
\centering
\vspace{-10pt}
\begin{tabular}{c}

\subfloat[\label{conv_pw_6_sample}]{\includegraphics[trim={0 0 0 32pt},clip, width=0.2\textwidth]{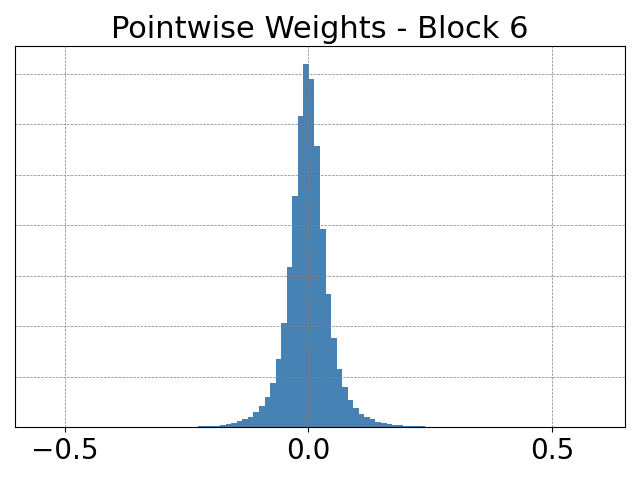}} 

\subfloat[\label{conv_dw_6_sample}]{\includegraphics[trim={0 0 0 32pt},clip, width=0.2\textwidth]{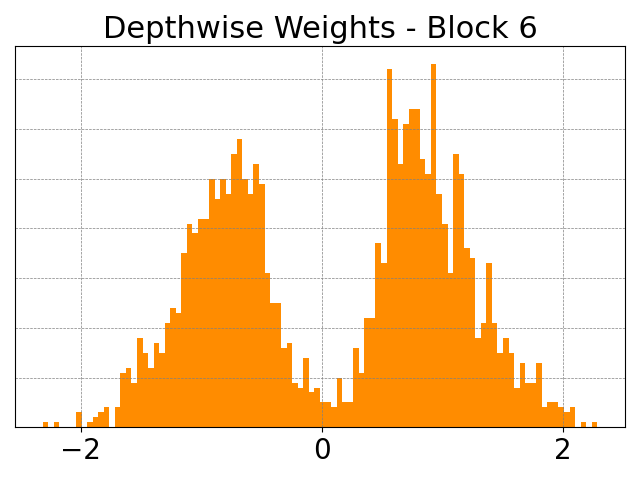}}
\end{tabular}
\vspace{-10pt}
\caption{Weights distribution of pre-trained PW and DW Convolution layers in PikeLPN where (a) Sample Pointwise layer weights (b) Sample Depthwise layer weights.}
\vspace{-15pt}
\label{fig:distribution_mismatch_dw_pw}
\end{figure}

\noindent \textbf{Quantizing Separable Convolution Layers:}
Linear quantizers results in a set of equally spaced values since they use affine mapping as shown in Equation \ref{eq:linear_quantizer}.
Non-uniform quantizers have different constraints. For example, Power-of-two (PoT) \cite{miyashita2016convolutional} restrict quantization levels to be powers-of-two values.
They can be used to increase the representational density of small values, furthermore, they have the benefit of replacing the multiplication operations during inference with shifts which are significantly cheaper as shown in Table \ref{tab:cost_metrics}.
However, using PoT quantizers for both pointwise (PW) and depthwise (DW) convolution operations in the separable convolution block leads to significant accuracy degradation as shown in the third row of Table \ref{tab:quantizers_accuracy_vs_energy}.
To get some insights, we analyze the distribution of the full-precision weights of PikeLPN when pre-trained on ImageNet.
Figures \ref{fig:distribution_mismatch_dw_pw}(a) and \ref{fig:distribution_mismatch_dw_pw}(b) visualize the distributions of a sample PW and DW weights respectively.
Interestingly, the majority of the weights of the PW layer lie around $\pm 0.1$, while the weights in the DW layer are distributed around $\pm 2$.
This mismatch in weights distribution across different layers makes low-precision quantization for the separable convolution blocks challenging because the used values fail to capture both distributions.
To address this problem, we propose using \textit{Distribution Heterogeneous Quantization} where the pointwise weights use the more efficient PoT quantizer while the depthwise weights use a linear quantizer.
It is important to note that pointwise convolutions contribute to 95\% of the number of multiply-accumulate operations in \textit{PikeLPN}; hence using the PoT quantizer in pointwise layers only improves the model's efficiency by 50\% as shown in Table \ref{tab:quantizers_accuracy_vs_energy}.

\noindent \textbf{Double Quantization:} Quantization requires extra elementwise multiplications by a floating-point scaling factor which add significant overhead as shown in Table \ref{tab:quantization_granularity}.
While we can not completely remove the scale factor, we can reduce the overhead from quantization scale multiplications by quantizing those quantization parameters.
We refer to quantizing the quantization parameters as \textit{Double Quantization}.
We consider using a PoT scale for the linear depthwise quantizer in \textit{PikeLPN} which can potentially reduce the elementwise operation from $3.7 mJ$ to $0.13 mJ$ based on Table \ref{tab:cost_metrics}.
Our experiments indicates negligible effect on accuracy when applying \textit{Double Quantization} as shown in Table \ref{tab:quantizers_accuracy_vs_energy}.

\noindent \textbf{Quantizing Batch Norm Layers:}
Batch normalization layers are used in most modern deep learning models to stabilize the training and improve their performance \cite{batchnorm}.
Batch normalization is computed as follows:
\vspace{-5pt}
\begin{equation}
    batchnorm(x) = \frac{(x-\mu)*\gamma}{\sqrt{\sigma^2+\epsilon}} + \beta
\label{bn}
\vspace{-5pt}
\end{equation}
Where $x$ is the input feature map and the batch norm parameters $\mu$, $\gamma$, $\sigma$, $\beta$ are represented as floating-point values.
To avoid performing floating point multiplications and additions, those parameters need to be quantized as follows:

\vspace{-10pt}
\begin{equation}
    Qbatchnorm(x) = \frac{(x-Q(\mu))*Q(\gamma)}{\sqrt{Q(\sigma)^2+\epsilon}} + Q(\beta)
\label{std_q_bn}
\vspace{-5pt}
\end{equation}

\begin{figure}[t]
  \centering
   \vspace{-10pt}
   \includegraphics[width=0.95\linewidth]{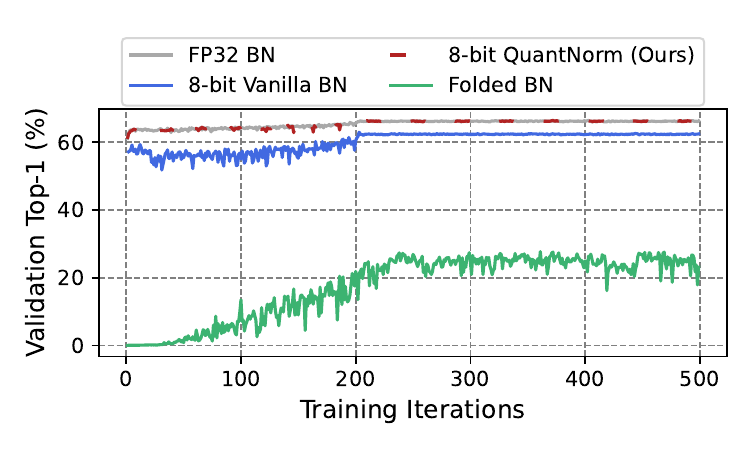}
   \vspace{-15pt}
   \caption{Validation Top-1 Accuracy during QAT on ImageNet for different Batch Norm Quantization techniques.}
   \label{fig:vanilla_bn_quantization}
   \vspace{-15pt}
\end{figure}

Computation folding is a commonly used approach to reduce the overhead of batch normalization operations in quantized models (i.e., mainly in 8 bit models) \cite{jacob2018quantization}. 
However, the batch normalization parameters (i.e., $\mu$, $\gamma$, $\sigma$, and $\beta$) have to be quantized to the same precision of the preceding convolution layers to enable folding.
Doing that in low-precision models (i.e., 4 bits or lower) leads to a significant loss in accuracy as shown in Figure \ref{fig:vanilla_bn_quantization}.
That is why previous low-precision model research \cite{zhang2022pokebnn, park2020profit, bin_mobilenet} excluded batch normalization layers from the quantization process, where they keep the batch norm parameters as floating point numbers.
However, as we showed earlier in Table \ref{tab:batchnorm_energy}, the non-quantized batch normalization operations can add up to 40\% overhead to the model's $ACE_{v2}$ cost.

Another solution is to quantize the batch normalization parameters at a higher precision.
Figure \ref{fig:vanilla_bn_quantization} shows the validation accuracy curve during training when batch normalization parameters are represented as INT8 values (denoted as \textit{8-bit Vanilla BN}).
Although the accuracy is better than the folded batch norm, we can still notice some degradation in accuracy compared to non-quantized batch norm layers.
To minimize the accuracy loss, we propose 
a novel \textit{QuantNorm} layer. In our  \textit{QuantNorm} layer, we re-write the batch norm quantization operation as shown in Equation \ref{improved_bn} where we first multiply by a quantized scale $s$, then add a quantized bias $b$.
$s$ is represented as the quantized division between the $\gamma$ and $\sigma$ parameters as shown in Equation \ref{bn_scale}.
Using \textit{QuantNorm} helps reduce quantization error by allowing high precision division in the scale $s$ computation during training. 
As shown in Figure \ref{fig:vanilla_bn_quantization}, our \textit{QuantNorm} layer maintains close-to-FP accuracy without any extra costs compared to vanilla quantization for batch norm layer. 
After training, we pre-compute $s$ to avoid high precision division during inference. 

\vspace{-10pt}
\begin{equation}
\label{improved_bn}
    Qbatchnorm(x)_{improved} = x*s - b   
\end{equation}
\vspace{-5pt}
\begin{equation}
\label{bn_scale}
    s = Q(\frac{\gamma}{\sqrt{\sigma^2+\epsilon}})
\end{equation}
\begin{equation}
\label{bn_bias}
    b = Q(\beta) - Q(\mu)*s
\end{equation}

\noindent \textbf{Model Scaling:}
To generate a Pareto family of models, we scale the number of output channels as practiced in the MobileNetV1 model \cite{howard2017mobilenets}. We also scale the precision of the input activation to the pointwise convolution layers in the PikeLPN block. 
We show more details about scaling \textit{PikeLPN} in Appendix Section \textcolor{red}{8}. 
\section{Experiments}
\label{sec:experiments}

\begin{table*}[t]
  \centering
  \setlength{\tabcolsep}{4pt}
\renewcommand{\arraystretch}{0.8}
\small
  \caption{Results \--- \textit{PikeLPN} versus SOTA low-precision models in terms of Accuracy and Efficiency Metrics. 
  $ACE_{v2}$ is measured according to the definition in Section \ref{extending_ace}. 
  The fourth and fifth columns show the contribution to the overall $ACE_{v2}$ cost by multiply-accumulate and elementwise operations respectively.
  \textit{Energy} represents the arithmetic energy according to $45nm$ CMOS technology according to table \ref{tab:cost_metrics}.
  \textit{Arithmetic Intensity} is an indication for the memory reads and writes required by the model as explained in Section \ref{key_bottlenecks}.
  \textit{Used Precisions} represent the the precision of the various operations in the mixed-precision models.
}
    \vspace{-5pt}
  \begin{tabular}{c | c | c | c | c | c | c | c}
    \toprule
     \multirow{2}{*}{\textbf{Model}} & \textbf{Accuracy} & \multicolumn{3}{c}{\textbf{Arithmetic Computational Effort ($\bm{ACE_{v2}}$)}} & \textbf{Energy} & \textbf{Arithmetic Intensity} & \textbf{Used} \\
    & (\%) &  Total ($\times10^9 \downarrow$) & MAC (\%) & Elementwise ($\%$) & ($mJ \downarrow$) & (Ops/Element $\uparrow$) & \textbf{Precisions} \\
    
    \midrule
    XNOR-Net \cite{rastegari2016xnor} & 51.2 & 143.78 & - & - & 587.69 & - & 32, 1 \\
    MobiNet \cite{phan2020mobinet} & 54.4 & 12.64 & 13.17 & 86.83 & 50.66 & 28 & - \\
    Bi-RealNet-18 \cite{liu2018bi} & 56.4 & 166.26 & - & - & 678.75 & - & 32, 1 \\
    Bi-RealNet-34 \cite{liu2018bi} & 62.2 & 168.11 & - & - & 691.47 & - & 32, 1 \\
    MobileNet (8W, 4A) \cite{krishnamoorthi2018quantizing} & 64.0 & 33.8 & 68.96 & 31.04 & 118.54 & 39.57 & 32, 8, 4 \\
    MobileNet (4W, 8A) \cite{krishnamoorthi2018quantizing} & 65.0 & 33.8 & 68.96 & 31.04 & 118.54 & 39.57 & 32, 8, 4 \\
    Real-to-Binary Net \cite{real_to_binary} & 65.4 &  186.85 & - & - & 762.24 & - & 32, 1 \\
    MeliusNet-29 \cite{bethge2020meliusnet} & 65.8 & 158.21 & - & - & 656.81 & - & 32, 1 \\
    PokeBNN-0.5x \cite{zhang2022pokebnn} & 65.2 & 33.58 & 4.18 & 95.81 & 143.78 & 24.5 & 32, 8, 4, 1 \\
    \midrule
   \textbf{PikeLPN-1$\times$} (Ours) & \textbf{67.55} & \textbf{8.50} & 96.38 & 3.62 & \textbf{34.98} & 39.57 & 8, 4 \\
    \midrule
    PROFIT \cite{park2020profit} & 69.05 & 20.91 & 47.51 & 52.49 & 82.70 & 39.57 & 32, 4 \\
    MeliusNet-42 \cite{bethge2020meliusnet} & 69.20 & 215.71 & - & - & 901.82 & - & 32, 1 \\
    \midrule
    \textbf{PikeLPN-2$\times$} (Ours) & \textbf{69.23} & \textbf{15.56} & 97.87 & 2.13 & \textbf{64.20} & 39.57 & 16, 8, 4\\
    \midrule
    ReActNet \cite{liu2020reactnet} & 69.4 & 83.24 & 26.78 & 73.22 & 361.63 & 36.75 & 32, 1 \\
    PokeBNN-0.75x \cite{zhang2022pokebnn} & 70.5 & 50.61 & 5.11 & 94.88 & 218.51 & 40.48 & 32, 8, 4, 1\\
    MobileNet (8bit) \cite{krishnamoorthi2018quantizing} & 70.7 &  51.44 & 79.61 & 20.39 & 173.68 & 39.57 & 32, 8 \\
    \midrule
    \textbf{PikeLPN-3$\times$} (Ours) & \textbf{71.95} &  \textbf{33.70} & 98.52 & 1.48 & \textbf{139.59} & \textbf{52.66} & 16, 8, 4 \\
    \midrule
    PokeBNN-1x \cite{zhang2022pokebnn} & 73.4 & 68.56 & 6.16 & 93.83 & 298.44 & 40.48 & 32, 8, 4, 1\\ 
    \midrule
    \textbf{PikeLPN-6$\times$} (Ours) & \textbf{73.59} &  \textbf{58.74} & 98.87 & 1.13 & \textbf{243.85} & \textbf{63.38} & 16, 8, 4 \\
    \bottomrule
  \end{tabular}
  \vspace{-15pt}
  \label{tab:comparison_to_sota}
\end{table*}

\subsection{Implementation and Training}
\label{sec:impl_training}
All models are implemented using QKeras \cite{coelho2021automatic}, then we performed Quantization-aware training (QAT) \cite{jacob2018quantization}.
We train and evaluate the \textit{PikeLPN} family of models on the ILSVRC12 ImageNet classification dataset \cite{imagenet}.
To train our low-precision models, we follow a multi-phase training approach.
We first train the full-precision model, then we quantize the model as explained previously in Subsection \ref{pikelpn}, and train for another $500$ epochs.
All Models are trained with an effective batch size of $256$ using an \textit{AdamW} optimizer and a Cosine Decay schedule.
We use label smoothing regularization with cross-entropy loss and a smoothing factor of $0.1$ for all models.
The initial learning rate is $1e-4$ and annealed using a cosine schedule to $1e-12$. 
An interesting observation was that training for the final $100$ epochs at a constant low learning-rate (i.e., $1e-12$) help the weights of the low-precision models stabilize and significantly boost the accuracy.
More details and visualization about this behaviour in added in the Appendix. 
We use standard augmentation techniques like resizing, cropping, and flipping. 
At test time, all \textit{PikeLPN} models are evaluated on images of resolution $224 \times 224$.

\begin{figure}[t]
  \centering
   \includegraphics[width=0.95\linewidth]{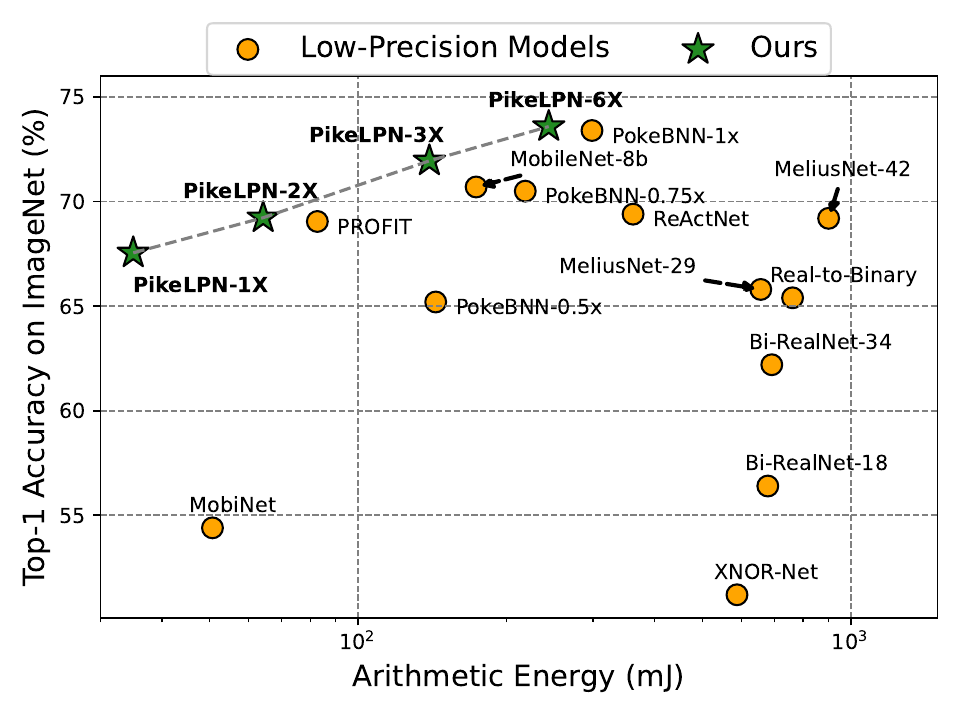}
   \vspace{-15pt}
   \caption{Accuracy and Energy Consumption by the arithmetic operations of our \textit{PikeLPN} vs SOTA low-precision neural networks.}
   \vspace{-17pt}
   \label{fig:sota_energy}
\end{figure}

\subsection{Results}
To evaluate the accuracy-efficiency trade-off by \textit{PikeLPN}, we compare its performance to state-of-the-art low-precision models.
Figures \ref{fig:sota_energy} and \ref{fig:sota_ace} show that \textit{PikeLPN} establishes the SOTA Pareto frontier for low-precision and binary models in terms of arithmetic energy consumption and $ACE_{v2}$ cost respectively.
Table \ref{tab:comparison_to_sota} compares \textit{PikeLPN} to SOTA low-precision models in terms of Top-1 Accuracy on ImageNet, Energy consumption in $millijoules$, $ACE_{v2}$, and Arithmetic Intensity.
We clearly see how the elementwise operations dominate (i.e., ~31-93\%) the $ACE_{v2}$ cost for other low-precision models.
On the other hand, \textit{PikeLPN} carefully quantizes the elementwise operations reducing their contribution to the total energy consumption to less than 5\%.
Additionally, \textit{PikeLPN-}$1\times$ is $1.5\times$ more efficient in terms of both $ACE_{v2}$ and arithmetic energy consumption compared to \textit{MobiNet} \cite{bin_mobilenet} (i.e., A binary version of MobileNetV1 with added skip connections) while achieving \textbf{13.2\%} higher Top-1 Accuracy on ImageNet.
Moreover, \textit{PikeLPN-}$3\times$ achieves $1.5\%$ higher Top-1 accuracy than PokeBNN-$0.75\times$ \cite{zhang2022pokebnn} (i.e., A binary ResNet-50 with parameterized activation functions) while being $35\%$ more efficient.
In terms of arithmetic intensity, \textit{PikeLPN} shows a much higher arithmetic intensity when compared to other low-precision models, this is mainly due to the absence of any skip connections.
As mentioned earlier in Section \ref{key_bottlenecks}, high arithmetic intensity is advantageous as it suggests a greater proportion of computational operations per data element, which can lead to reducing the memory reads and writes by the model; hence reducing the overall energy consumption during inference.
\section{Conclusion}
\label{conclusion}
Our investigation into SOTA low-precision models uncovered overlooked efficiency bottlenecks, particularly noting that operations traditionally considered negligible—such as elementwise operations in activation functions, batch normalization, and quantization scaling can contribute up to 90\% of the inference cost.
Addressing these challenges, we proposed $ACE_{v2}$ which extends the efficiency metric \textit{ACE} to better reflect the inference cost of low-precision models.
Moreover, we introduced \textit{PikeLPN}, a novel family of models that quantizes both elementwise and multiply-accumulate operations.
Specifically, we propose (a) a novel \textit{QuantNorm} layer for effective batch normalization quantization, (b) \textit{Double Quantization} where quantization parameters are also quantized, and (c) \textit{Distribution-Heterogeneous Quantization} for Separable Convolution layers to tackle their distribution mismatch problem.
\textit{PikeLPN} achieves up to a threefold reduction in inference cost over existing low-precision models while improving the Top-1 accuracy in ImageNet dataset.


{
    \small
    \bibliographystyle{ieeenat_fullname}
    \bibliography{main}
}
\clearpage
\setcounter{page}{1}
\maketitlesupplementary

\section{Elementwise Multiplications}
\label{sec:fp_ele_muls}
In Section \textcolor{red}{3.2}, we propose $ACE_{v2}$ which extends ACE to account for elementwise multiplications.
We derived the number of adders required for multiplying an \( i \)-bit number by a \( j \)-bit number as \( i \cdot j - \max(i, j) \).
Here we provide a more detailed justification for the derived formula.
For simplicity, we assume both operands have the same number of bits (i.e., $i$ = $j$) in this derivation.
Elementwise multiplications requires a multiplier as well as an adder to account for the dot pattern at the completion of the multiplication \cite{wallace_and_dadda}.
We base our derivation on the established implementation of Dadda multiplier \cite{wallace_and_dadda} and Ripple-Carry Adder (RCA) \cite{rca} to estimate the cost of elementwise multiplications.
To multiply two $i$-bit numbers, the Dadda multiplier requires \(i^2 - 3i + 2 \) adders \cite{wallace_and_dadda}, while the RCA adds another \(2i - 2 \) adders. 
This leads to a total number of adders equal to \(i^2 - i \).
To generalize to operands with different precisions, the cost for elementwise multiplications between an \( i \)-bit number and a \( j \)-bit number can be derived as \( i \cdot j - \max(i, j) \), with \( i \cdot j \) reflecting the cost of the multiplier and \( \max(i, j) \) representing the final addition. 
Independently, we performed an empirical verification for \( 1 \leq i, j \leq 64 \) which confirmed the correctness of this formula, showing zero error in predicted adder counts. 
This refinement in $ACE_{v2}$ cost calculation enhances our understanding of multiplier complexity.

\section{Floating Point Elementwise Additions}
\label{sec:fp_ele_adds}
In Section \textcolor{red}{3.2}, we improve ACE by extending it to include the cost of floating-point elementwise addition.
We derive the cost of adding an $i$-bit and a $j$-bit floating point numbers, using the formula 
\begin{equation}
\label{ace_add_fp}
    ACE_{fp-add} = c_a \cdot \max(i, j)
\end{equation}
For simplicity, we assume both operands have the same number of bits (i.e., $i$ = $j$) in this derivation.
$c_a$ reflects the added complexity of floating point operations compared to fixed-point addition.
To derive $c_a$, we look into the components of floating-point adders \cite{fp_adders} and analyze the $ACE_{v2}$ cost for each component. 
Assuming $e$ bits for the exponent and $m$ bits for the mantissa, the main components of the floating-point adder and their corresponding $ACE_{v2}$ costs are as follows:

\begin{enumerate}
\item \textit{Exponent Subtraction}: Involves subtracting the exponent bits resulting in an $ACE_{v2}$ cost of $\mathbf{e}$.
\item \textit{Operand Swapping}: Requires a single multiplexer with negligible $ACE_{v2}$ cost.
\item \textit{Limitation of Alignment Shift Amount}: Involves adding the mantissa bits resulting in an $ACE_{v2}$ cost of $\mathbf{m}$.
\item \textit{Alignment Shift}: Involves shifting by the mantissa bits adding an $ACE_{v2}$ cost of $\mathbf{m \cdot log_2(m)/5}$\footnote{$ACE_{v2}$ cost for shift operation is derived as $i \cdot \log_2(j)/5$ in Subsection $3.2$}.
\item \textit{Significand Negation}: Involves one bit subtraction resulting in an $ACE_{v2}$ cost of \textbf{1}.
\item \textit{Significand Addition}: Requires mantissa bits addition resulting in an $ACE_{v2}$ cost of $\mathbf{m}$.
\item \textit{Significand Conversion}: Requires two additions adding an $ACE_{v2}$ cost of $\mathbf{2m}$.
\item \textit{Normalization}: Requires shifting $e$ bits resulting in an $ACE_{v2}$ cost of $\mathbf{e \cdot log_2(e)/5}$.
\item \textit{Rounding and Post-normalization}: Requires adding $m$ bits with an $ACE_{v2}$ cost of $\mathbf{m}$.
\end{enumerate}

Summing the costs for all the components, we get a total cost of $m(5 + \log_2(m)/5) + e + e \cdot \log_2(e)/5 + 1$. 
Considering the dominant role of mantissa operations, we approximate the total cost to $i (5 + log_2(i)/5)$ where $i$ is the number of bits of the added floating point number. 
The upper bound for $\log_2(i)/5$ is $1$ when m is $32$.
Therefore, we can derive the cost as $6 \dot i$ resulting in $c_a = 6$ in Equation \ref{ace_add_fp}. 
This approximation streamlines $ACE_{v2}$ calculation for floating-point additions.
To verify its correctness, we show that it aligns well with the independently measured energy consumption observed in 45nm CMOS technology in Table \textcolor{red}{1}.

\section{Model Scaling}
\label{sec:model_scaling}
To scale PikeLPN-1$\times$ to the 2$\times$, 3$\times$, and 6$\times$ sizes, we employ a series of scaling techniques including multiplying the output channels of the convolution layers by a scaling factor and increasing the precision of the feature maps at the point-wise convolution layers. 
These techniques increase both the $ACE_{v2}$ cost and the representational capacity of the model, allowing us to generate a Pareto family of models. 
The details for each model are described in Table \ref{tab:pikelpn_variants}.
For nomenclature, the scale factor represents the $ACE_{v2}$ cost of the scaled model compared to that of the smallest model. 
For example, PikeLPN-$3\times$ has approximately 3 times the $ACE_{v2}$ cost of PikeLPN-$1\times$.

\begin{table}[t]
  \centering
  \setlength{\tabcolsep}{2.5pt}
  \renewcommand{\arraystretch}{1.2}
  \caption{Comparison of PikeLPN variants' training parameters, with dropout rates calibrated to mitigate overfitting. Each model's training duration and learning rate strategy are customized according to its complexity. They are initialized with weights from an floating point PikeLPN model, employing a consistent learning rate of $10^{-12}$ during the tail period to enhance stability and validation accuracy, crucial for smaller models.}
  \vspace{-5pt}
  \small
  \begin{tabular}{c | c | c | c | c}
    \hline
    \textbf{PikeLPN Size}& \textbf{1$\times$} & \textbf{2$\times$} & \textbf{3$\times$} & \textbf{6$\times$} \\
    \hline
    $\bm{ACE_{v2}}$ ($\times 10^9$) & 8.68 & 15.74 & 33.97 & 59.10 \\
    \hline
    \textbf{Channel Multiplier} & 1.0 & 1.0 & 1.5 & 2.0 \\
    \hline
    \textbf{Activation precision} & (6, 1, 1) & (8, 7, 1) & (8, 7, 1) & (8, 7, 1) \\
    (int bits, frac bits, sign bit) & & & & \\
    \hline
    \textbf{Removal of BN layers } & Yes & No & No & No \\
    between depthwise and & & & & \\
    pointwise convolutions& & & & \\
    \hline
    \textbf{Constant learning rate} & 300 & 300 & 20 & 50 \\
    \textbf{tail period} (epochs)& & & & \\
    \hline
    \textbf{Training epochs} & 500 & 1500 & 1000 & 1000 \\
    \hline
    \textbf{Dropout rate} & 1e-3 & 1e-3 & 0.5 & 0.7 \\
    \hline
  \end{tabular}
  \label{tab:pikelpn_variants}
\end{table}

\begin{figure}[t]
  \centering
  \vspace{-10pt}
   \includegraphics[width=0.95\linewidth]{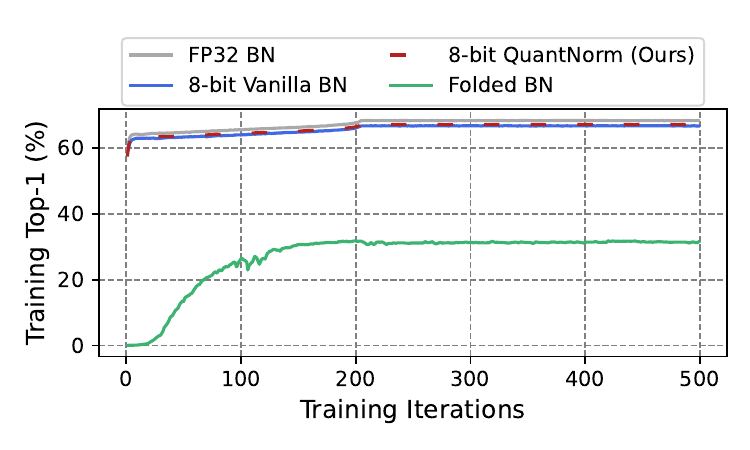}
   \vspace{-10pt}
   \caption{Training Top-1 Accuracy during QAT on ImageNet at different Batch Norm Quantization techniques for PikeLPN-$1\times$.}
   \vspace{-20pt}
   \label{fig:vanilla_bn_quantization_train_acc}
\end{figure} 

\begin{figure}
  \centering
  \begin{subfigure}{\linewidth}
   \includegraphics[width=\linewidth]{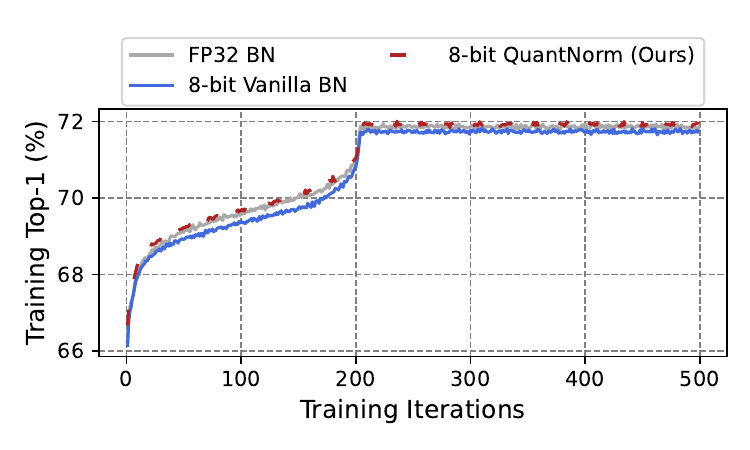} 
   \caption{Training Accuracy (\%)}
    \label{fig:vanilla16_bn_quantization_train}
   \includegraphics[width=\linewidth]{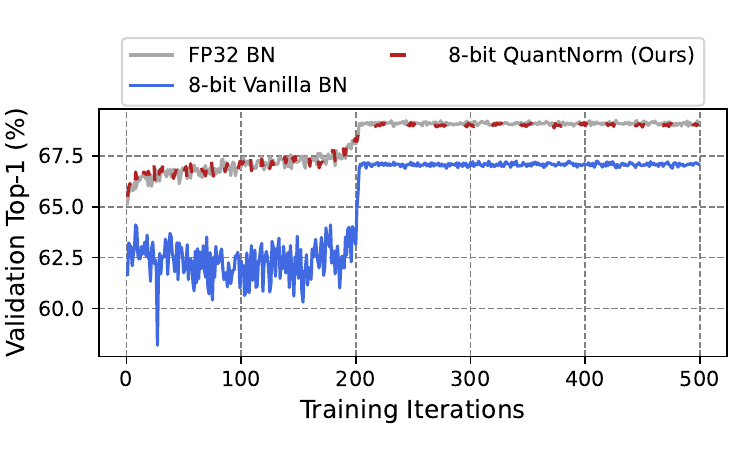}
   \caption{Validation Accuracy (\%)}
   \label{fig:vanilla16_bn_quantization_val}
  \end{subfigure}
  \caption{Top-1 Accuracy during QAT on ImageNet at different Batch Norm Quantization techniques for PikeLPN-$2\times$.}
  \label{fig:vanilla16_bn_quantization}

  \end{figure}

\section{QuantNorm Layer}
\label{sec:bn_quantization}
As shown in Subsection \textcolor{red}{3.3}, \textit{QuantNorm} reduces quantization error during training improving the performance of our \textit{PikeLPN} models on ImageNet image classification dataset \cite{imagenet}.
As shown in Figure \textcolor{red}{$6$}, \textit{QuantNorm} maintains close-to-FP validation accuracy when using it during PikeLPN-$1\times$ training. 
Figure \ref{fig:vanilla_bn_quantization_train_acc} shows the top-1 training accuracy while training the same model using different batch normalization quantization techniques.
Moreover, to ensure that the same behaviour persists at different \textit{PikeLPN} scales, Figures \ref{fig:vanilla16_bn_quantization_train} and \ref{fig:vanilla16_bn_quantization_val} shows the top-1 training and validation accuracies respectively of PikeLPN-$2\times$ when using our proposed \textit{QuantNorm} layer versus the vanilla batch norm quantization shown in Equation \textcolor{red}{5}.

\begin{figure}[t]
  \centering
   \includegraphics[width=0.95\linewidth]{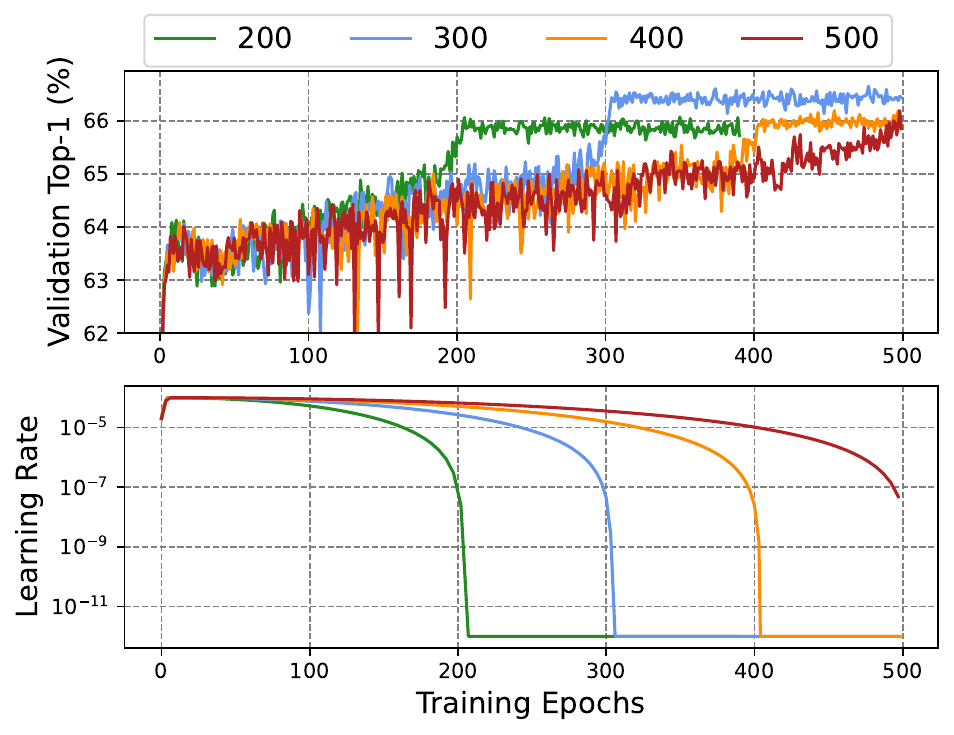}
   \vspace{-10pt}
   \caption{Validation Top-1 Accuracy for PikeLPN-1$\times$ model with various learning rate decay schedules. All the training sessions match exactly except for the number of decay steps which ranges between 200 and 500 epochs.}
   \label{fig:lr_decay}
   \vspace{-10pt}
\end{figure}

\section{Early LR Decay}
\label{sec:early_lr_decay}
As mentioned in Subsection \textcolor{red}{4.1}, our \textit{PikeLPN} models are trained using an \textit{AdamW} optimizer and a Cosine Decay Schedule.
The initial learning rate is $1e-4$ and annealed using a cosine schedule to $1e-12$.
Figure \ref{fig:lr_decay} summarizes the training behaviour with various learning rate schedules.
The x-axis represents the training iterations which we limit to 500 epochs.
The y-axis in the top graph represents the validation top-1 accuracy on ImageNet, while the y-axis in the bottom graph represents the learning rate.
All the training sessions match exactly except for the number of decay steps which ranges between 200 and 500 epochs.
The figure highlights two main observations.
First, training for the final few epochs at a constant low learning-rate (i.e., $1e-12$) help the weights of the low-precision models stabilize and significantly boost the accuracy (i.e., $1-2\%$).
Second, the number of decay steps is an important hyper-parameter when training low-precision models.
For example, we noticed that for PikeLPN-1$\times$, setting the number of decay steps to 300 gives an extra $0.5-1\%$ improvement in validation accuracy.

\begin{table}[t]
  \centering
  \setlength{\tabcolsep}{2.5pt}
\renewcommand{\arraystretch}{1}
\small
  \caption{\textit{PikeLPN} versus baselines \-- detailed analysis for contribution of elementwise operations to $ACE_{v2}$.
  \textit{Total} represents the total percentage of elementwise operations from $ACE_{v2}$. \textit{BN}, \textit{ACT} and \textit{QP} represents the detailed contribution of batch norm layers, activation layers and quantization overhead respectively.
}
  \begin{tabular}{c | c | c | c c c }
    \toprule
     \multirow{2}{*}{\textbf{Model}} & \multirow{2}{*}{\textbf{MAC $\bm{ACE_{v2}}$}} & \multicolumn{4}{c}{\textbf{Elementwise $\bm{ACE_{v2}}$}} \\
    & & Total & BN & Act & QP \\
    \midrule
    PokeBNN-0.5x & 4.2 & 95.8\% & 43.4\% &  29.2\% & 21.5\% \\
   \textbf{PikeLPN-1$\times$} & 96.4 & 3.9\% & 3.7\% & 0\% & 0.2\% \\
    \midrule
    PROFIT & 48\% & 52\% & 19\% & 17\% & 16\% \\
    \textbf{PikeLPN-2$\times$} & 97.9 & 2.1\% & 2\% & 0\% & 0.1\% \\
    \midrule
    PokeBNN-1x & 6.2 & 93.8\% & 42.2\% &  28.4\% & 20.9\% \\
   \textbf{PikeLPN-6$\times$} & 98.9 & 1.13\% & 0.98\% & 0\% & 0.2\% \\
    \bottomrule
  \end{tabular}
  \vspace{-15pt}
  \label{tab:sample_results}
\end{table}

\section{Detailed Analysis for Contribution of Elementwise operations to \texorpdfstring{$\bm{ACE_{v2}}$}{ACEv2}}
Table \ref{tab:sample_results} shows the detailed contributions of different elementwise computation sources to the overall $ACE_{v2}$ cost.

\section{Comparing ACE to similar metrics}
\label{other_metrics_comparison}
In Section \textcolor{red}{3.2}, we propose an extension to ACE, but our proposed extension could in principle generalize to other metrics similar to ACE. All previous metrics similar to ACE known by the authors only account for accumulate/dot-product operations and not elementwise operations, making our proposed extension generally valuable. Even so, we chose to specifically extend the ACE metric as opposed to other metrics due to ACE's simplicity and efficacy in predicting energy costs. We extend ACE because it was built with ML researchers in mind, creating balance between complexity and abstraction of hardware energy which \-- from a physics perspective in CMOS \-- is likely to limit future ML hardware.
We would like to compare ACE to four other metrics that are often brought up when trying to predict hardware costs: (1) Fanout-of-4 inverter delay (FO4) \cite{sutherland1999logical} is a constraint in hardware design, but not necessarily a target for ML researchers. (2) Ristretto \cite{gysel2018ristretto} measures power cost through a labor-intensive synthesis, inaccessible to ML researchers. (3, 4) The Unit-gate model \cite{zimmermann1999computer} and full-adder count \cite{sakr2017analytical} correlate very well with ACE and differ only by a small constant factor \footnote{Unlike $ACE_{v2}$, their application to DNNs does not take the cost elementwise operations into account nor does it account for carry-save format for local accumulator representations typically used in systollic arrays.}.
Therefore our $ACE_{v2}$ extension would generalize to these metrics.
Moreover, all those metrics, including $ACE_{v2}$, are technology independent.
\end{document}